\documentclass{article}

\usepackage[margin=1in]{geometry}
\usepackage{url,hyperref,lineno,microtype,subcaption}
\usepackage{siunitx,graphicx}

\title{Reinforcement Learning\\with Low-Complexity Liquid State Machines}

\author{
    Wachirawit Ponghiran\thanks{Equal contributors to this work}~~~~~Gopalakrishnan Srinivasan$^{*}$~~~~~Kaushik Roy\\
    Department of ECE\\
    Purdue University\\
    West Lafayette, IN 47906\\
  \texttt{wponghir@purdue.edu} \\
}

\begin{document}

\maketitle

\begin{abstract}
We propose reinforcement learning on simple networks consisting of random connections of spiking neurons (both recurrent and feed-forward) that can learn complex tasks with very little trainable parameters. Such sparse and randomly interconnected recurrent spiking networks exhibit highly non-linear dynamics that transform the inputs into rich high-dimensional representations based on past context. The random input representations can be efficiently interpreted by an output (or readout) layer with trainable parameters. Systematic initialization of the random connections and training of the readout layer using Q-learning algorithm enable such small random spiking networks to learn optimally and achieve the same learning efficiency as humans on complex reinforcement learning tasks like Atari games. The spike-based approach using small random recurrent networks provides a computationally efficient alternative to state-of-the-art deep reinforcement learning networks with several layers of trainable parameters. The low-complexity spiking networks can lead to improved energy efficiency in event-driven neuromorphic hardware for complex reinforcement learning tasks.
\end{abstract}
\section{Introduction}

High degree of recurrent connectivity among neuronal populations is a key attribute of neural microcircuits in the cerebral cortex and many different brain regions \cite{douglas1995recurrent,harris2013cortical,jiang2015principles}. Such common structure suggests the existence of a general principle for information processing. However, the principle underlying information processing in such recurrent population of spiking neurons is still largely elusive due to the complexity of training large recurrent Spiking Neural Networks (SNNs). In this regard, reservoir computing architectures \cite{maass2002real, maass2003model, lukovsevivcius2009reservoir} were proposed to minimize the training complexity of large recurrent neuronal populations. Liquid State Machine (LSM) \cite{maass2002real, maass2003model} is a recurrent SNN consisting of an input layer sparsely connected to a randomly interlinked reservoir (or liquid) of excitatory and inhibitory spiking neurons whose activations are passed on to a readout (or output) layer, trained using supervised algorithms, for inference. The key attribute of an LSM is that the input-to-liquid and the recurrent excitatory$\leftrightarrow$inhibitory synaptic connectivity matrices and weights are fixed \textit{a priori}. LSM effectively utilizes the rich nonlinear dynamics of Leaky-Integrate-and-Fire spiking neurons \cite{dayan2003theoretical} and the sparse random input-to-liquid and recurrent-liquid synaptic connectivity for processing spatio-temporal inputs. At any time instant, the spatio-temporal inputs are transformed into a high-dimensional representation, referred to as the liquid states (or spike patterns), which evolves dynamically based on decaying memory of the past inputs. The memory capacity of the liquid is dictated by its size and degree of recurrent connectivity. Although the LSM, by construction, does not have stable instantaneous internal states like Turing machines \cite{savage1998models} or attractor neural networks \cite{amit1992modeling}, prior studies have successfully trained the readout layer using liquid activations, estimated by integrating the liquid states (spikes) over time, for speech recognition \cite{maass2002real, auer2002reducing, verstraeten2005isolated, bellec2018long}, image recognition \cite{srinivasan2018spilinc}, gesture recognition \cite{chrolcannon2015learning, panda2018learning}, and sequence generation tasks \cite{panda2017learning, nicola2017supervised, bellec2019biologically}.

In this work, we propose such sparse randomly-interlinked low-complexity LSMs for solving complex Reinforcement Learning (RL) tasks, which involve an autonomous agent (modeled using the LSM) trained to select actions in a manner that maximizes the expected future rewards received from the environment. For instance, a robot (agent) learning to navigate a maze (environment) based on the reward and punishment received from the environment is an example RL task. At any given time, the environment state (converted to spike trains) is fed to the liquid, which produces a high-dimensional liquid state (spike pattern) based on decaying memory of the past environment states. We present an optimal initialization strategy for the fixed input-to-liquid and recurrent-liquid synaptic connectivity matrices and weights to enable the liquid to produce high-dimensional representations that lead to efficient training of the liquid-to-readout weights. Artificial rate-based neurons for the readout layer takes the liquid activations and produces \textit{action-values} to guide action selection for a given environment state. The liquid-to-readout weights are trained using the Q-learning RL algorithm proposed for deep learning networks \cite{mnih2015human}. In RL theory \cite{sutton1998reinforcement}, the Q-value, also known as the action-value, estimates the expected future rewards for a state-action pair that specifies how good is the action for the current environment state. The readout layer of the LSM contains as many neurons as the number of possible actions for a particular RL task. At any given time, the readout neurons predict the Q-value for all possible actions based on the high-dimensional state representation provided by the liquid. The liquid-to-readout weights are then trained using backpropagation \cite{rumelhart1986learning} to minimize the error between the Q-values predicted by the LSM and the target Q-values estimated from RL theory \cite{watkins1992q} as described in \autoref{sec:LSM_training}. We adopt $\epsilon$-greedy policy (explained in \autoref{sec:LSM_training}) to select the appropriate action based on the predicted Q-values during training and evaluation. Based on $\epsilon$-greedy policy, a lot of random actions are picked in the beginning of the training phase to better explore the environment. Towards the end of training and during inference, the action corresponding to the maximum Q-value is selected with higher probability to exploit the learnt experiences. We first demonstrate results for training the readout weights based on the high-dimensional representations provided by the liquid, as a result of the sparse recurrent-liquid connectivity, on simple Cartpole-balancing RL task \cite{sutton1998reinforcement}. We then comprehensively validate the capability of the LSM and the presented training methodology on complex RL tasks like Pacman \cite{denero2016pacman} and Atari games \cite{brockman2016openai}. We note that LSM has been previously trained using Q-learning for RL tasks pertaining to robotic motion control \cite{joshi2005movement, berberich2017implementation, tieck2018learning}. We demonstrate and benchmark the efficacy of appropriately initialized LSM for solving RL tasks commonly used to evaluate deep reinforcement learning networks. In essence, this work provides a promising step towards incorporating bio-plausible low-complexity recurrent SNNs like LSMs for complex RL tasks, which can potentially lead to much improved energy efficiency in event-driven asynchronous neuromorphic hardware implementations \cite{merolla2014million, davies2018loihi}.

\section{Introduction}

High degree of recurrent connectivity among neuronal populations is a key attribute of neural microcircuits in the cerebral cortex and many different brain regions \cite{douglas1995recurrent,harris2013cortical,jiang2015principles}. Such common structure suggests the existence of a general principle for information processing. However, the principle underlying information processing in such recurrent population of spiking neurons is still largely elusive due to the complexity of training large recurrent Spiking Neural Networks (SNNs). In this regard, reservoir computing architectures \cite{maass2002real, maass2003model, lukovsevivcius2009reservoir} were proposed to minimize the training complexity of large recurrent neuronal populations. Liquid State Machine (LSM) \cite{maass2002real, maass2003model} is a recurrent SNN consisting of an input layer sparsely connected to a randomly interlinked reservoir (or liquid) of excitatory and inhibitory spiking neurons whose activations are passed on to a readout (or output) layer, trained using supervised algorithms, for inference. The key attribute of an LSM is that the input-to-liquid and the recurrent excitatory$\leftrightarrow$inhibitory synaptic connectivity matrices and weights are fixed \textit{a priori}. LSM effectively utilizes the rich nonlinear dynamics of Leaky-Integrate-and-Fire spiking neurons \cite{dayan2003theoretical} and the sparse random input-to-liquid and recurrent-liquid synaptic connectivity for processing spatio-temporal inputs. At any time instant, the spatio-temporal inputs are transformed into a high-dimensional representation, referred to as the liquid states (or spike patterns), which evolves dynamically based on decaying memory of the past inputs. The memory capacity of the liquid is dictated by its size and degree of recurrent connectivity. Although the LSM, by construction, does not have stable instantaneous internal states like Turing machines \cite{savage1998models} or attractor neural networks \cite{amit1992modeling}, prior studies have successfully trained the readout layer using liquid activations, estimated by integrating the liquid states (spikes) over time, for speech recognition \cite{maass2002real, auer2002reducing, verstraeten2005isolated, bellec2018long}, image recognition \cite{srinivasan2018spilinc}, gesture recognition \cite{chrolcannon2015learning, panda2018learning}, and sequence generation tasks \cite{panda2017learning, nicola2017supervised, bellec2019biologically}.

In this work, we propose such sparse randomly-interlinked low-complexity LSMs for solving complex Reinforcement Learning (RL) tasks, which involve an autonomous agent (modeled using the LSM) trained to select actions in a manner that maximizes the expected future rewards received from the environment. For instance, a robot (agent) learning to navigate a maze (environment) based on the reward and punishment received from the environment is an example RL task. At any given time, the environment state (converted to spike trains) is fed to the liquid, which produces a high-dimensional liquid state (spike pattern) based on decaying memory of the past environment states. We present an optimal initialization strategy for the fixed input-to-liquid and recurrent-liquid synaptic connectivity matrices and weights to enable the liquid to produce high-dimensional representations that lead to efficient training of the liquid-to-readout weights. Artificial rate-based neurons for the readout layer takes the liquid activations and produces \textit{action-values} to guide action selection for a given environment state. The liquid-to-readout weights are trained using the Q-learning RL algorithm proposed for deep learning networks \cite{mnih2015human}. In RL theory \cite{sutton1998reinforcement}, the Q-value, also known as the action-value, estimates the expected future rewards for a state-action pair that specifies how good is the action for the current environment state. The readout layer of the LSM contains as many neurons as the number of possible actions for a particular RL task. At any given time, the readout neurons predict the Q-value for all possible actions based on the high-dimensional state representation provided by the liquid. The liquid-to-readout weights are then trained using backpropagation \cite{rumelhart1986learning} to minimize the error between the Q-values predicted by the LSM and the target Q-values estimated from RL theory \cite{watkins1992q} as described in \autoref{sec:LSM_training}. We adopt $\epsilon$-greedy policy (explained in \autoref{sec:LSM_training}) to select the appropriate action based on the predicted Q-values during training and evaluation. Based on $\epsilon$-greedy policy, a lot of random actions are picked in the beginning of the training phase to better explore the environment. Towards the end of training and during inference, the action corresponding to the maximum Q-value is selected with higher probability to exploit the learnt experiences. We first demonstrate results for training the readout weights based on the high-dimensional representations provided by the liquid, as a result of the sparse recurrent-liquid connectivity, on simple Cartpole-balancing RL task \cite{sutton1998reinforcement}. We then comprehensively validate the capability of the LSM and the presented training methodology on complex RL tasks like Pacman \cite{denero2016pacman} and Atari games \cite{brockman2016openai}. We note that LSM has been previously trained using Q-learning for RL tasks pertaining to robotic motion control \cite{joshi2005movement, berberich2017implementation, tieck2018learning}. We demonstrate and benchmark the efficacy of appropriately initialized LSM for solving RL tasks commonly used to evaluate deep reinforcement learning networks. In essence, this work provides a promising step towards incorporating bio-plausible low-complexity recurrent SNNs like LSMs for complex RL tasks, which can potentially lead to much improved energy efficiency in event-driven asynchronous neuromorphic hardware implementations \cite{merolla2014million, davies2018loihi}.


\section{Materials and Methods} \label{sec:methods}

\subsection{Liquid State Machine: Architecture and Initialization} \label{sec:LSM_arch_init}

\begin{figure}[!htbp]
\begin{center}
\includegraphics[width=0.90\columnwidth]{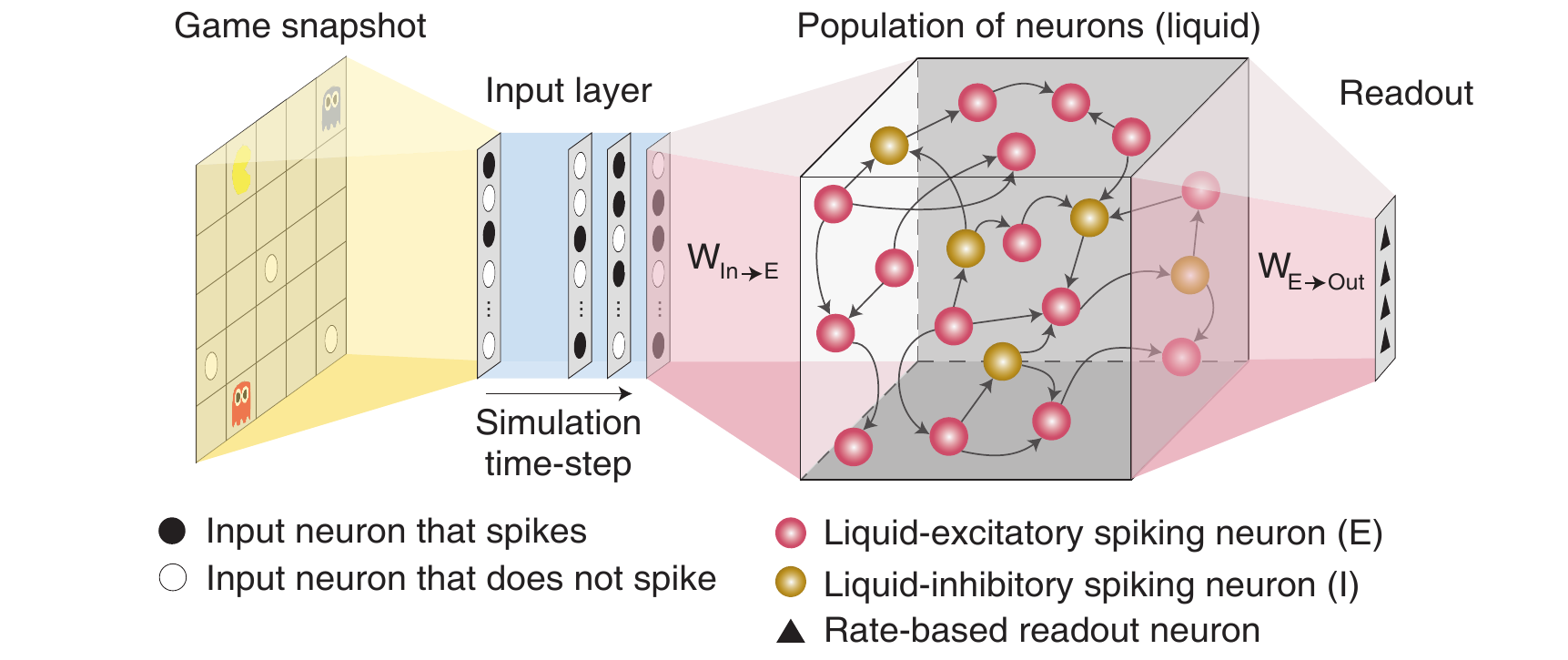}
\end{center}
\caption{Illustration of the LSM architecture consisting of an input layer sparsely connected via fixed synaptic weights to randomly recurrently connected reservoir (or liquid) of excitatory and inhibitory spiking neurons followed by a readout layer composed of artificial rate-based neurons.}
\label{fig:intro_lsm}
\end{figure}

Liquid State Machine (LSM) consists of an input layer sparsely connected via fixed synaptic weights to a randomly interlinked liquid of excitatory and inhibitory spiking neurons followed by a readout layer as depicted in \textbf{\autoref{fig:intro_lsm}}. The input layer (denoted by $P$) is modeled as a group of excitatory neurons that spike based on the input environment state following a Poisson process. The sparse input-to-liquid connections are initialized such that each excitatory neuron in the liquid receives synaptic connections from approximately $K$ random input neurons. This guarantees uniform excitation of the liquid-excitatory neurons by the external input spikes. The fixed input-to-liquid synaptic weights are chosen from a uniform distribution between 0 and $\alpha$ as shown in \autoref{tab:weight}, where $\alpha$ is the maximum bound imposed on the weights. The liquid consists of excitatory neurons (denoted by $E$) and inhibitory neurons (denoted by $I$) recurrently connected in a sparse random manner as illustrated in  \textbf{\autoref{fig:intro_lsm}}. The number of excitatory neurons is chosen to be $4\times$ the number of inhibitory neurons as observed in the cortical circuits \cite{wehr2003balanced}. We use the Leaky-Integrate-and-Fire (LIF) model \cite{dayan2003theoretical} to mimic the  dynamics of both excitatory and inhibitory spiking neurons as described by the following differential equations:

\begin{equation}
    \frac{dV_{i}}{dt} = \frac{V_{rest} - V_{i}}{\tau} + I_{i}(t)\label{eq:vmem}
\end{equation}

\begin{equation}
I_{i}(t) = \sum_{{l}\in{N_{P}}} W_{li}~{\cdot}~\delta(t-t_l) + \sum_{{j}\in{N_{E}}} W_{ji}~{\cdot}~\delta(t-t_j) - \sum_{{k}\in{N_{I}}}W_{ki}~{\cdot}~\delta(t-t_k)    \label{eq:cur}
\end{equation}
where $V_{i}$ is the membrane potential of the $i$-th neuron in the liquid, $V_{rest}$ is the resting potential to which $V_{i}$ decays to, with time constant $\tau$, in the absence of input current, and $I_{i}(t)$ is the instantaneous current projecting into the $i$-th neuron, and $N_{P}$, $N_{E}$, and $N_{I}$ are the number of input, excitatory, and inhibitory neurons, respectively. The instantaneous current is a sum of three terms: current from input neurons, current from excitatory neurons, and current from inhibitory neurons. The first term integrates the sum of pre-synaptic spikes, denoted by $\delta(t-t_l)$ where $t_l$ is the time instant of pre-spikes, with the corresponding synaptic weights ($W_{li}$ in \autoref{eq:cur}). Likewise, the second (third) term integrates the sum of pre-synaptic spikes from the excitatory (inhibitory) neurons, denoted by $\delta(t-t_j)$ ($\delta(t-t_k)$), with the respective weights $W_{ji}$ ($W_{ki}$) in \autoref{eq:cur}. The neuronal membrane potential is updated with the sum of the input, excitatory, and negative inhibitory currents as shown in \autoref{eq:vmem}. When the membrane potential reaches a certain threshold $V_{thres}$, the neuron fires an output spike. The membrane potential is thereafter reset to $V_{reset}$ and the neuron is restrained from spiking for an ensuing refractory period by holding its membrane potential constant. The LIF model parameters for the excitatory and inhibitory neurons are listed in \autoref{tab:neuron_param}.

There are four types of recurrent synaptic connections in the liquid, namely, $E{\rightarrow}E$, $E{\rightarrow}I$, $I{\rightarrow}E$, and $I{\rightarrow}I$. We express each connection in the form of a matrix that is initialized to be sparse and random, which causes the spiking dynamics of a particular neuron to be independent of most other neurons and maintains separability in the neuronal spiking activity. However, the degree of sparsity needs to be tuned to achieve rich network dynamics. We find that excessive sparsity (reduced connectivity) leads to weakened interaction between the liquid neurons and renders the liquid memoryless. On the contrary, lower sparsity (increased connectivity) results in chaotic spiking activity, which eliminates the separability in neuronal spiking activity. We initialize the connectivity matrices such that each excitatory neuron receives approximately $C$ synaptic connections from inhibitory neurons, and vice versa. The hyperparameter $C$ is tuned empirically as discussed in~\autoref{sec:finding_lsminit} to avoid common chaotic spiking activity problems that occur when (1) excitatory neurons connect to each other and form a loop that always leads to positive drift in membrane potential, and when (2) an excitatory neuron connects to itself and repeatedly gets excited from its activity.
Specifically, for the first situation, we have non-zero elements in the connectivity matrix $E{\rightarrow}E$ (denoted by $W_{EE}$) only at locations where elements in the product of connectivity matrices $E{\rightarrow}I$ and $I{\rightarrow}E$ (denoted by $W_{EI}$ and $W_{IE}$, respectively) are non-zero. This ensures that excitatory synaptic connections are created only for those neurons that also receive inhibitory synaptic connections, which mitigates the possibility of continuous positive drift in the respective membrane potentials. To circumvent the second situation, we force the diagonal elements of $W_{EE}$ to be zero and eliminate the possibility of repeated self-excitation. Throughout this work, we create a recurrent connectivity matrix for liquid with $m$ excitatory neurons and $n$ inhibitory neurons by forming an $m \times n$ matrix whose values are randomly drawn from a uniform distribution between $0$ and $1$. Connection is formed between those pairs of neurons where the corresponding matrix entries are lesser than the target connection probability ($=C/m$). For illustration, consider a liquid with $m{=}1000$ excitatory and $n{=}250$ inhibitory neurons. In order to create the $E{\rightarrow}I$ connectivity matrix such that each inhibitory neuron receives synaptic connection from a single excitatory neuron ($C{=}1$), we first form a $1000\times250$ random matrix whose values are drawn from a uniform distribution between $0$ and $1$. We then create a connection between those pairs of neurons where the matrix entries are lesser than 0.1\% (1/1000). Similar process is repeated for connection $I{\rightarrow}E$. We then initialize connection $E{\rightarrow}E$ based on the product of $W_{EI}$ and $W_{IE}$. Similarly, the connectivity matrix for $I{\rightarrow}I$ (denoted by $W_{II}$) is initialized based on the product of $W_{IE}$ and $W_{EI}$. The connection weights are initialized from a uniform distribution between $0$ and $\beta$ as shown in \autoref{tab:weight} for different recurrent connectivity matrices. Note that the weights of the synaptic connections from inhibitory neurons are greater than that for synaptic connections from excitatory neurons to account for the lower number of inhibitory neurons relative to excitatory neurons. Stronger inhibitory connection weights help ensure that every neuron receives similar amount of excitatory and inhibitory input currents, which improves the stability of the liquid as experimentally validated in \autoref{sec:finding_lsminit}.

The liquid-excitatory neurons are fully-connected to artificial rate-based neurons in the readout layer for inference. The readout layer, which consists of as many output neurons as the number of actions for a given RL task, uses the average firing rate/activation of the excitatory neurons to predict the Q-value for every state-action pair. We translate the liquid spiking activity to average rate by accumulating the excitatory neuronal spikes over the time period for which the input (current environment state) is presented. We then normalize the spike counts with the maximum possible spike count over the LSM-simulation period, which is computed as the LSM-simulation period divided by the simulation time-step, to obtain the average firing rate of the excitatory neurons that are fed to the readout layer. Since the number of excitatory neurons is larger than the number of output neurons in the readout layer, we gradually reduce the dimension by introducing an additional fully-connected hidden layer between the liquid and the output layer. We use ReLU non-linearity \cite{nair2010rectified} after the first hidden layer but none after the final output layer since the Q-values are unbounded and can assume positive or negative values. We train the synaptic weights constituting the fully-connected readout layer using the Q-learning based training methodology that is described in the following \autoref{sec:LSM_training}.

\subsection{Q-Learning Based LSM Training Methodology} \label{sec:LSM_training}

Reinforcement Learning (RL) tasks fundamentally involve an agent (for instance, a robot) that is trained to navigate a certain environment (for instance, a maze) in a manner that maximizes the total rewards in the future. Formally, at any time instant $t$, the agent receives the environment state $s_t$ and picks action $a_t$ from the set of all possible actions. After the environment receives the action $a_t$, it transitions to the next state based on the chosen action and feeds back an immediate reward $r_{t+1}$ and the new environment state $s_{t+1}$. As mentioned in the beginning, the goal of the agent is to maximize the accumulated reward in the future, which is mathematically expressed as

\begin{equation}
    R_t = \sum_{t=1}^{\infty} \gamma^{t}~ r_{t}
\end{equation}
where $\gamma \in [0, 1]$ is the discount factor that determines the relative significance attributed to immediate and future reward. If $\gamma$ is chosen to be 0, the agent maximizes only the immediate reward. However, as $\gamma$ approaches unity, the agent learns to maximize the accumulated reward in the future. Q-learning \cite{watkins1992q} is a widely used RL algorithm that enables the agent to achieve this objective by computing the state-action value function (or commonly known as the Q-function), which is the expected future reward for a state-action pair that is specified by

\begin{equation}
Q_{\pi}(s,a) = \mathrm{E}[R_t| s_t=s, a_t=a, \pi]     \label{eq:q_1}
\end{equation}
where $Q_{\pi}(s,a)$ measures the value of choosing an action $a$ when in state $s$ following a policy $\pi$. If the agent follows the optimal policy (denoted by $\pi_\ast$) such that $Q_{\pi_\ast}(s,a) = \max\limits_{\pi} Q_{\pi}(s,a)$, the Q-function can be estimated recursively using the Bellman optimality equation that is described by

\begin{equation}
Q_{\pi_\ast}(s,a) = \mathrm{E}[r_{t+1}+\gamma \max_{a_{t+1}} Q_{\pi_\ast} (s_{t+1},a_{t+1})|s,a]    \label{eq:q_2}
\end{equation}
where $Q_{\pi_\ast}(s,a)$ is the Q-value for choosing action $a$ from state $s$ following the optimal policy $\pi_\ast$, $r_{t+1}$ is the immediate reward received from the environment, $Q_{\pi_\ast}(s_{t+1},a_{t+1})$ is the Q-value for selecting action $a_{t+1}$ from the next environment state $s_{t+1}$. Learning the Q-values for all possible state-action pairs is intractable for practical RL applications. Popular approaches approximate Q-function using deep convolutional neural networks \cite{mnih2015human, mnih2016asynchronous, lillicrap2015continuous, silver2016mastering}.

In this work, we model the agent using an LSM, wherein the liquid-to-readout weights are trained to approximate the Q-function as described below. At any time instant $t$, we map the current environment state vector $s_t$ to input neurons firing at a rate constrained between $0$ and $\phi$~\si{\hertz} over certain time period (denoted by $T_{LSM}$) following a Poisson process. The maximum Poisson firing rate $\phi$ is tuned to ensure sufficient input spiking activity for a given RL task. We follow the method outlined in \cite{heeger2000poisson} to generate the Poisson spike trains as explained below. For a particular input neuron in the state vector, we first compute the probability of generating a spike at every LSM-simulation time-step based on the corresponding Poisson firing rate. Note that the time-steps in the RL task are orthogonal to the time-steps used for the numerical simulation of the liquid. Specifically, in-between successive time-steps $t$ and $t+1$ in the RL task, the liquid is simulated for a time period of $T_{LSM}$ with $1$\si{ms} separation between consecutive LSM-simulation time-steps. The probability of producing a spike at any LSM-simulation time-step is obtained by scaling the corresponding firing rate by $1{,}000$. We generate a random number drawn from a uniform distribution between $0$ and $1$, and produce a spike if the random number is lesser than the neuronal spiking probability. At every LSM-simulation time-step, we feed the spike map of the current environment state and record the spiking outputs of the liquid-excitatory neurons. We accumulate the excitatory neuronal spikes and normalize the individual neuronal spike counts with the maximum possible spike count over the LSM-simulation period to obtain the high-dimensional representation (activation) of the environment state as discussed in the previous \autoref{sec:LSM_arch_init}. It is important to note that appropriate initialization of the LSM (detailed in \autoref{sec:LSM_arch_init}) is necessary to obtain useful high-dimensional representation for efficient training of the liquid-to-readout weights as experimentally validated in \autoref{sec:finding}.

The high-dimensional liquid activations are fed to the readout layer that is trained using backpropagation to approximate the Q-function by minimizing the mean square error between the Q-values predicted by the readout layer and the target Q-values following \cite{mnih2015human} as described by the following equations:

\begin{equation}
\theta_{t+1} = \theta_{t}+\eta\left( Y_{t}-Q(s_{t},a_{t}|\theta_{t})\right) \nabla_{\theta_{t}} Q(s_{t},a_{t}|\theta_{t})    \label{eq:loss_1}
\end{equation}
\begin{equation}
Y_{t} = r_{t+1} + \gamma \max_{a_{t+1}} Q (s_{t+1},a_{t+1}|\theta_{t})    \label{eq:loss_2}
\end{equation}
where $\theta_{t+1}$ and $\theta_{t}$ are the updated and previous synaptic weights in the readout layer, respectively, $\eta$ is learning rate, $Q(s_{t},a_{t}|\theta_{t})$ is vector representing the Q-values predicted by the readout layer for all possible actions given the current environment state $s_t$ using the previous readout weights, $\nabla_{\theta_{t}} Q(s_{t},a_{t}|\theta_{t})$ is the gradient of the Q-values with respect to the readout weights, and $Y_{t}$ is the vector containing the target Q-values that is obtained by feeding the next environment state $s_{t+1}$ to the LSM while using the previous readout weights. To encourage exploration during training, we follow $\epsilon$-greedy policy \cite{watkins1989learning} for selecting the actions based on the Q-values predicted by the LSM. Based on $\epsilon$-greedy policy, we select a random action with probability $\epsilon$ and the optimal action, i.e., the action pertaining to the highest Q-value with probability $(1{-}\epsilon)$ during training. Initially, $\epsilon$ is set to a large value (closer to unity), thereby permitting the agent to pick a lot of random actions and effectively explore the environment. As training progresses, $\epsilon$ gradually decays to a small value, thereby allowing the agent to exploit its past experiences. During evaluation, we similarly follow $\epsilon$-greedy policy albeit with much smaller $\epsilon$ so that there is a strong bias towards exploitation. Employing $\epsilon$-greedy policy during evaluation also serves to mitigate the negative impact of over-fitting or under-fitting. In an effort to further improve stability during training and achieve better generalization performance, we use the experience replay technique proposed by \cite{mnih2015human}. Based on experience replay, we store the experience discovered at each time-step (i.e. $s_t$, $a_t$, $r_t$, and $s_{t+1}$) in a large table and later train the LSM by sampling mini-batches of experiences in a random manner over multiple training epochs, leading to improved generalization performance. For all the experiments reported in this work, we use the RMSProp algorithm \cite{tieleman2012lecture} as the optimizer for error backpropagation with mini-batch size of $32$. We adopt $\epsilon$-greedy policy, wherein $\epsilon$ gradually decays from $1$ to $0.001{-}0.1$ over the first $10\%$ of the training steps. Replay memory stores one million recently played frames, which are then used for mini-batch weight updates that are carried out after the initial $100$ training steps. The simulation parameters for Q-learning are summarized in \autoref{tab:qlearning_param}.


\section{Experimental Results} \label{sec:finding}

We first present results motivating the importance of careful LSM initialization for obtaining rich high-dimensional state representation, which is necessary for efficient training of the liquid-to-readout weights.
We then demonstrate the utility of the recurrent-liquid synaptic connections of careful LSM initialization using classic cartpole-balancing RL task \cite{sutton1998reinforcement}. We then validate the capability of appropriately initialized LSM, trained using the presented methodology, for solving complex RL tasks like Pacman \cite{denero2016pacman} and Atari games \cite{brockman2016openai}.

\subsection{LSM Hyperparameter Tuning} \label{sec:finding_lsminit}

Initializing LSM with appropriate parameters is an important step to construct a model that produces useful high-dimensional representations. Since the input-to-liquid and recurrent-liquid connectivity matrices of the LSM are fixed \textit{a priori} during training, how these connections are initialized dictates the liquid dynamics. We choose the parameters $K$ (governing the input-to-liquid connectivity matrix) and $C$ (governing the recurrent-liquid connectivity matrices) empirically based on three observations: (1) stable spiking activity of the liquid, (2) eigenvalue analysis of the recurrent connectivity matrices, and (3) development of liquid-excitatory neuron membrane potential.

\begin{figure}[!htbp]
\begin{center}
\includegraphics[width=0.85\columnwidth]{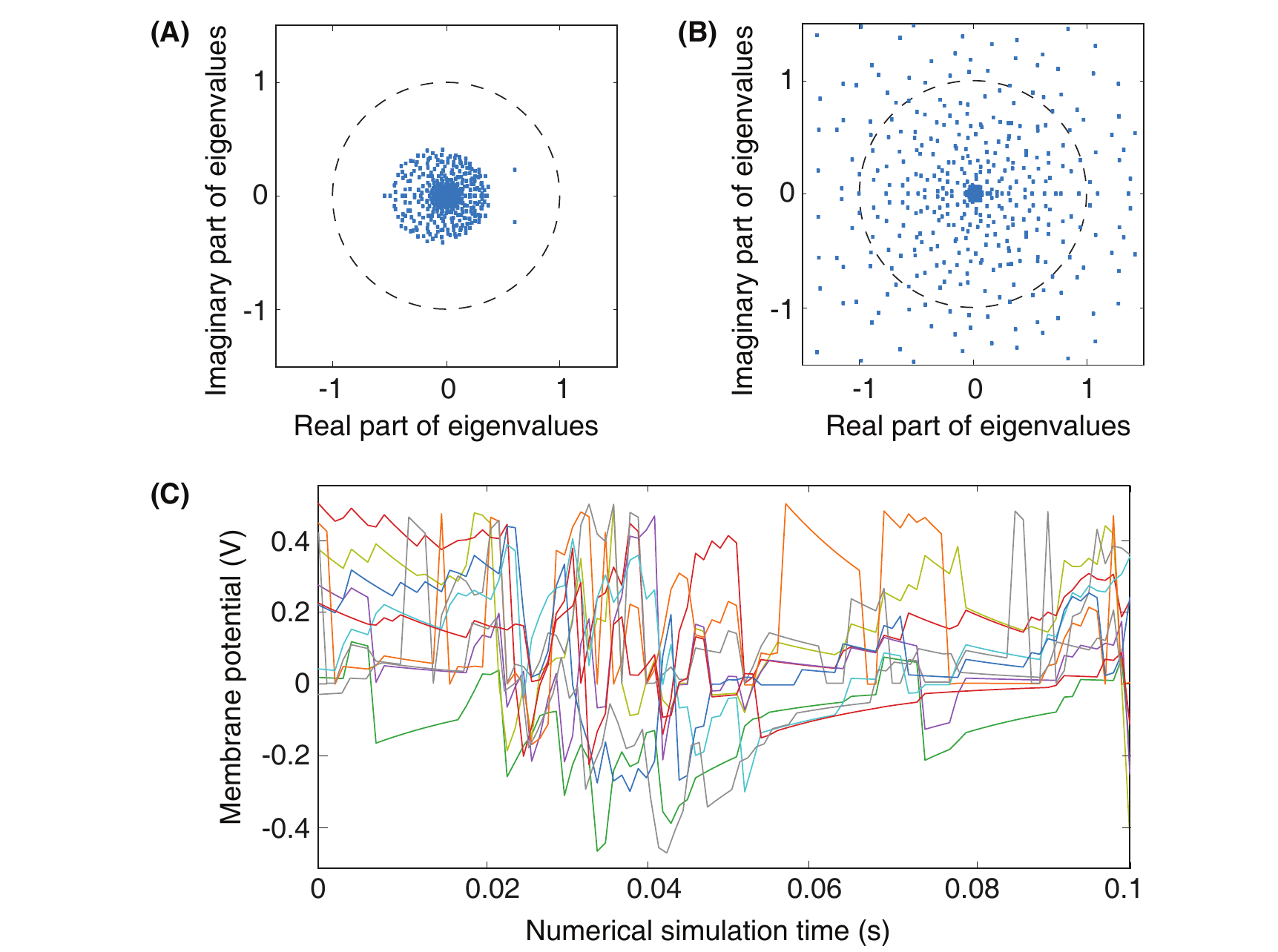}
\end{center}
\caption{\textbf{Metrics for guiding hyperparameter tuning: (A)} Eigenvalue spectrum of the recurrent-liquid connectivity matrix for an LSM containing $500$ liquid neurons. The LSM is initialized with synaptic weights listed in~\autoref{tab:weight} based on hyperparameter $C{=}4$. All eigenvalues in the spectrum lie within a unit circle. \textbf{(B)}  Eigenvalue spectrum of the recurrent-liquid connectivity matrix initialized with synaptic weights $\beta_{E{\rightarrow}E}=0.4$, $\beta_{E{\rightarrow}I}=0.1$, and $\beta_{I{\rightarrow}E}=0.1$. Many eigenvalues in the spectrum are outside the unit circle. \textbf{(C)} Development of membrane potentials from 10 randomly picked excitatory neurons in the liquid initialized with  synaptic weights listed in~\autoref{tab:weight} based on hyperparameter $C{=}4$. Random representation from the cartpole-balancing problem is used as the input.}
\label{fig:result_tune}
\end{figure}

Spiking activity of the liquid is said to be stable if every finite stream of inputs results in a finite period of response. Sustained activity indicates that small input noise can perturb the liquid state and lead to chaotic activity that is no longer dependent on the input stimuli. It is impractical to analyze the stability of the liquid for all possible input streams within a finite time. We investigate the liquid stability by feeding in random input stimuli and sampling the excitatory neuronal spike counts at regular time intervals over the LSM-simulation period for different values of $K$ and $C$. We separately adjust these parameters for each learning task using random representations of the environment from the games. Values of $K$ and $C$ are experimentally determined to be $3$ and $4$ for cartpole and Pacman experiment, respectively, which ensures stable liquid spiking activity while enabling the liquid to exhibit fading memory of the past inputs. Fading memory indicates that the liquid retains input information for a short period of time after the input stimuli are cut-off.

Analyzing the eigenvalue spectrum of the recurrent connectivity matrix is another tool to assess the stability of the liquid. Each eigenvalue in the spectrum represents an individual mode of the liquid. Real part indicates decay rate of the mode while the imaginary part corresponds to the frequency of the mode \cite{rajan2010stimulus}. Liquid spiking activity remains stable as long as all eigenvalues remain within the unit circle. However, this condition is not easily met for realistic recurrent-liquid connections with random synaptic weight initialization~\cite{rajan2006eigenvalue}. We constrain the recurrent weights (hyperparameter $\beta$) such that each neuron receives balanced excitatory and inhibitory synaptic currents as previously discussed in~\autoref{sec:LSM_arch_init}. This results in eigenvalues that lie within the unit circle as illustrated in \textbf{\autoref{fig:result_tune}(A)}. In order to emphasize the importance of LSM initialization, we also show the eigenvalue spectrum of the recurrent-liquid connectivity matrix when the weights are not properly initialized as shown in \textbf{\autoref{fig:result_tune}(B)} where many eigenvalues are outside the unit circle. Finally, we also use the development of the excitatory neuronal membrane potential to guide hyperparameter tuning. The hyperparameters $C$ and $\beta$ are chosen to ensure that membrane potential exhibits balanced fluctuation as illustrated in \textbf{\autoref{fig:result_tune}(C)} that plots the membrane potential of 10 randomly picked neurons in the liquid.

\subsection{Learning to Balance a Cartpole}

\begin{figure}[!htbp]
\begin{center}
\includegraphics[width=0.85\columnwidth]{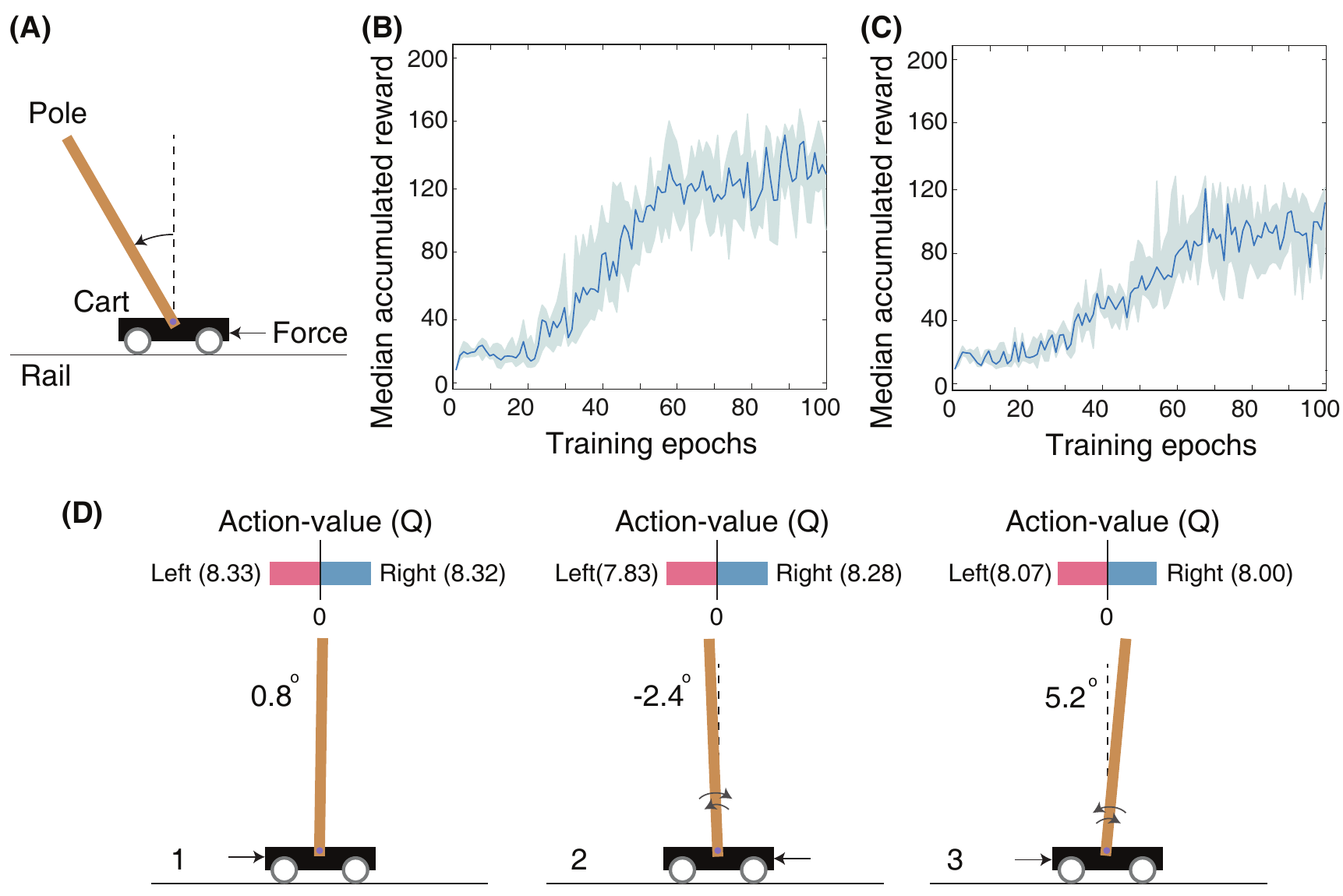}
\end{center}
\caption{\textbf{(A)} Illustration of the cartpole-balancing task wherein the agent has to balance a pole attached to a wheeled cart that moves freely on a rail of certain length. \textbf{(B)} The median accumulated reward per epoch provided by the LSM trained across $10$ different random seeds for the cartpole-balancing task. Shaded region in the plot represents the $25$-th to $75$-th percentile of the accumulated reward over multiple random seeds. \textbf{(C)} The median accumulated reward per epoch from cartpole training across $10$ different random seeds in which the LSM is initialized to have sparser connectivity between the liquid neurons compared to that used for the experiment in \textbf{(B)}.  \textbf{(D)} Visualization of the learnt Q (action-value) function for the cartpole-balancing task at three different game-steps designated as $1$, $2$, and $3$. Angle of the pole is written on the left side of each figure. Negative angle represents an unbalanced pole to the left and positive angle represents an unbalanced pole to the right. Black arrow corresponds to a unit force on the left or right side of the cart depending on which Q value is larger.}
\label{fig:result_learning_cartpole}
\end{figure}

Cartpole-balancing is a classic control problem wherein the agent has to balance a pole attached to a wheeled cart that can move freely on a rail of certain length as shown in \textbf{\autoref{fig:result_learning_cartpole}(A)}. The agent can exert a unit force on the cart either to the left or right side for balancing the pole and keeping the cart within the rail. The environment state is characterized by cart position, cart velocity, angle of the pole, and angular velocity of the pole, which are designated by the tuple $({\chi},\dot{\chi},{\varphi},\dot{\varphi})$. The environment returns a unit reward every time-step and concludes after $200$ time-steps if the pole does not fall or the cart does not goes out of the rail. Because the game is played for a finite time period, we constrain $({\chi},\dot{\chi},{\varphi},\dot{\varphi})$ to be within the range specified by $(\pm 2.5, \pm 0.5, \pm 0.28, \pm 0.88)$ for efficiently mapping the real-valued state inputs to spike trains feeding into the LSM. Each real-valued state input is mapped to $10$ input neurons which have firing rates proportional to one-hot encoding of the input value representing $10$ distinct levels within the corresponding range.

We model the agent using an LSM containing $150$ liquid neurons, $32$ hidden neurons in the fully-connected layer between the liquid and output layer, and $2$ output neurons. The maximum firing rate for the input neurons representing the environment state is set to $100$~\si{\hertz}. The LSM is trained for $10^{5}$ time-steps, which are equally divided into $100$ training epochs containing $1{,}000$ time-steps per epoch. After each epoch, the LSM is evaluated for $1{,}000$ time-steps with the probability of choosing a random action $\epsilon$ set to $0.05$. Note that the LSM is evaluated for $1{,}000$ time-steps (multiple gameplays) even though single gameplay lasts a maximum of only $200$ time-steps as mentioned in the previous paragraph. We use the accumulated reward averaged over multiple gameplays as the true indicator of the LSM (agent) performance to account for the randomness in action-selection as a result of the $\epsilon$-greedy policy. We train the LSM initialized with $10$ different random seeds and obtain median accumulated reward as shown in \textbf{\autoref{fig:result_learning_cartpole}(B)}. Note that the maximum possible accumulated reward per gameplay is $200$ since each gameplay lasts at most $200$ time-steps. Increase in median accumulated reward over epochs indicates that the LSM learnt to balance the cartpole using the dynamically evolving high-dimensional liquid states. The ability of the liquid to provide rich high-dimensional input representations can be attributed to the careful initialization of the connectivity matrices and weights (explained in \autoref{sec:LSM_arch_init}), which ensures balance between the excitatory and inhibitory currents to the liquid neurons and preserves fading memory of past liquid activity. However, the median accumulated reward after $100$ training epochs saturates around $125$ and does not reach the maximum value of $200$. We hypothesize that the game score saturation comes from the quantized representation of the environment state, and demonstrate in the following experiment with Pacman that the LSM can learn optimally given a better state representation. Finally, in order to emphasize the importance of LSM initialization, we also show the median accumulated reward per training epoch for training in which the LSM is initialized to have few synaptic connections. \textbf{\autoref{fig:result_learning_cartpole}(C)} indicates that the median accumulated reward is around $90$ when the LSM initialization is suboptimal.

To visualize the learnt action-value function guiding action selection, we compare Q-values produced by the LSM during evaluation in three different scenarios depicted in \textbf{\autoref{fig:result_learning_cartpole}(D)}. Note that each Q-value represents how good is the corresponding action for a given environment state. In scenario $1$ (see \textbf{\autoref{fig:result_learning_cartpole}(D)-1}) that corresponds to the beginning of the gameplay wherein the pole is almost balanced, the value of both the actions are identical. This implies that either action (moving the cart left or right) will lead to a similar outcome. In scenario $2$ (see \textbf{\autoref{fig:result_learning_cartpole}(D)-2}) wherein the pole is unbalanced to the left side, the difference between the predicted Q values increases. Specifically, the Q value for applying a unit force on the right side of the cart is higher, which causes the cart to move to the left. Pushing the cart to the left in turn causes the pole to swing back right towards the balanced position. Similarly, in scenario $3$ (see \textbf{\autoref{fig:result_learning_cartpole}(D)-3}) wherein the pole is unbalanced to the right side, the Q value is higher for applying a unit force on the left side of the cart, which causes the cart to move right and enables the pole to swing left towards the balanced position. This visually demonstrates the ability of the LSM (agent) to successfully balance the pole by pushing the cart appropriately to the left or right based on the learnt Q values.

\subsection{Learning to Play Pacman}

\begin{figure}[!htbp]
\begin{center}
\includegraphics[width=0.85\columnwidth]{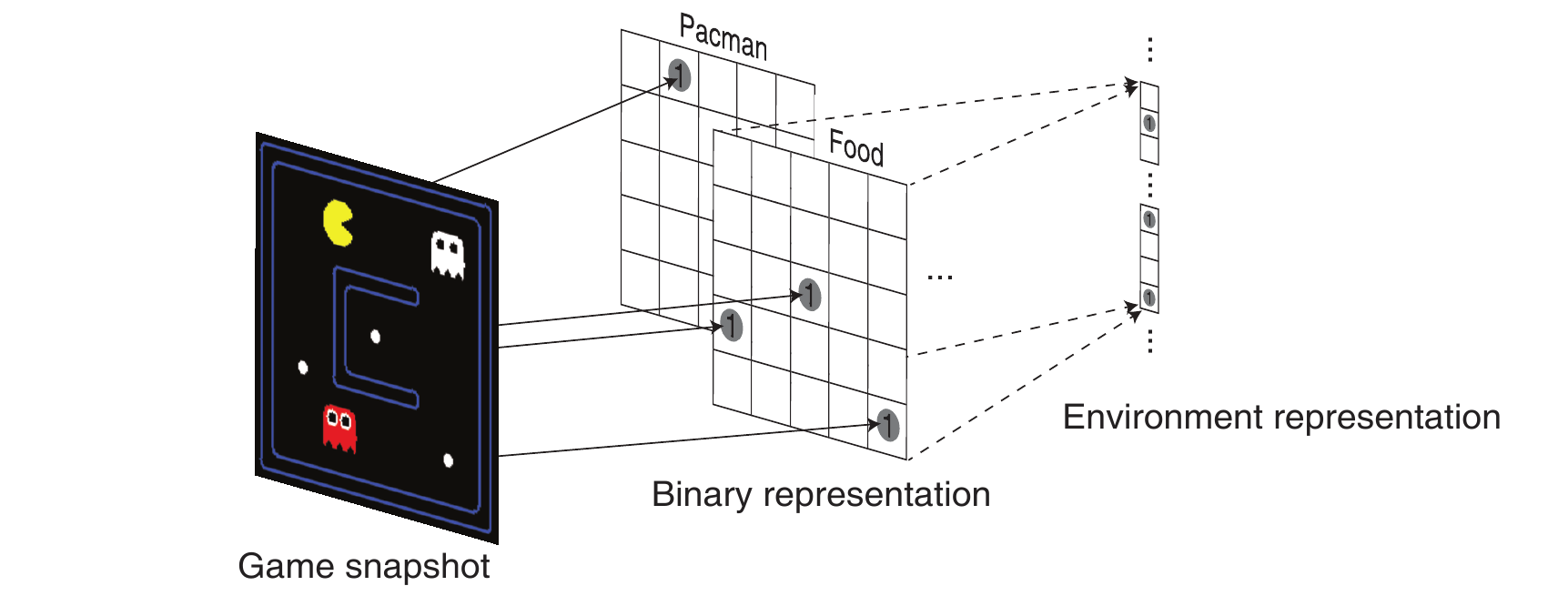}
\end{center}
\caption{Illustration of a snapshot from the Pacman game that is translated into $5$ two-dimensional binary representations corresponding to the location of Pacman, foods, cherries, ghosts, and scared ghosts. The binary intermediate representations are then flattened and concatenated to obtain the environment state representation.}
\label{fig:result_repr_pacman}
\end{figure}

In order to comprehensively validate the efficacy of the high-dimensional environment representations provided by the liquid, we train the LSM to play a game of Pacman \cite{denero2016pacman}. The objective of the game is to control Pacman (yellow in color) to capture all the foods (represented by small white dots) in a grid without being eaten by the ghosts as illustrated in \textbf{\autoref{fig:result_repr_pacman}}. The ghosts always hunt the Pacman; however, cherry (represented by large white dots) make the ghosts temporarily scared of the Pacman and run away. The game environment returns unit reward whenever Pacman consumes food, cherry, or the scared ghost (white in color). The game environment also returns a unit reward and restarts when all foods are captured. We use the location of Pacman, food, cherry, ghost and scared ghost as the environment state representation. The location of each object is encoded as a two-dimensional binary array whose dimension matches with that of the Pacman grid as shown in \textbf{\autoref{fig:result_repr_pacman}}. The binary intermediate representations of all the objects are then concatenated and flattened into a single vector to be fed to the input layer of the LSM.

\begin{figure}[!htbp]
\begin{center}
\includegraphics[width=0.85\columnwidth]{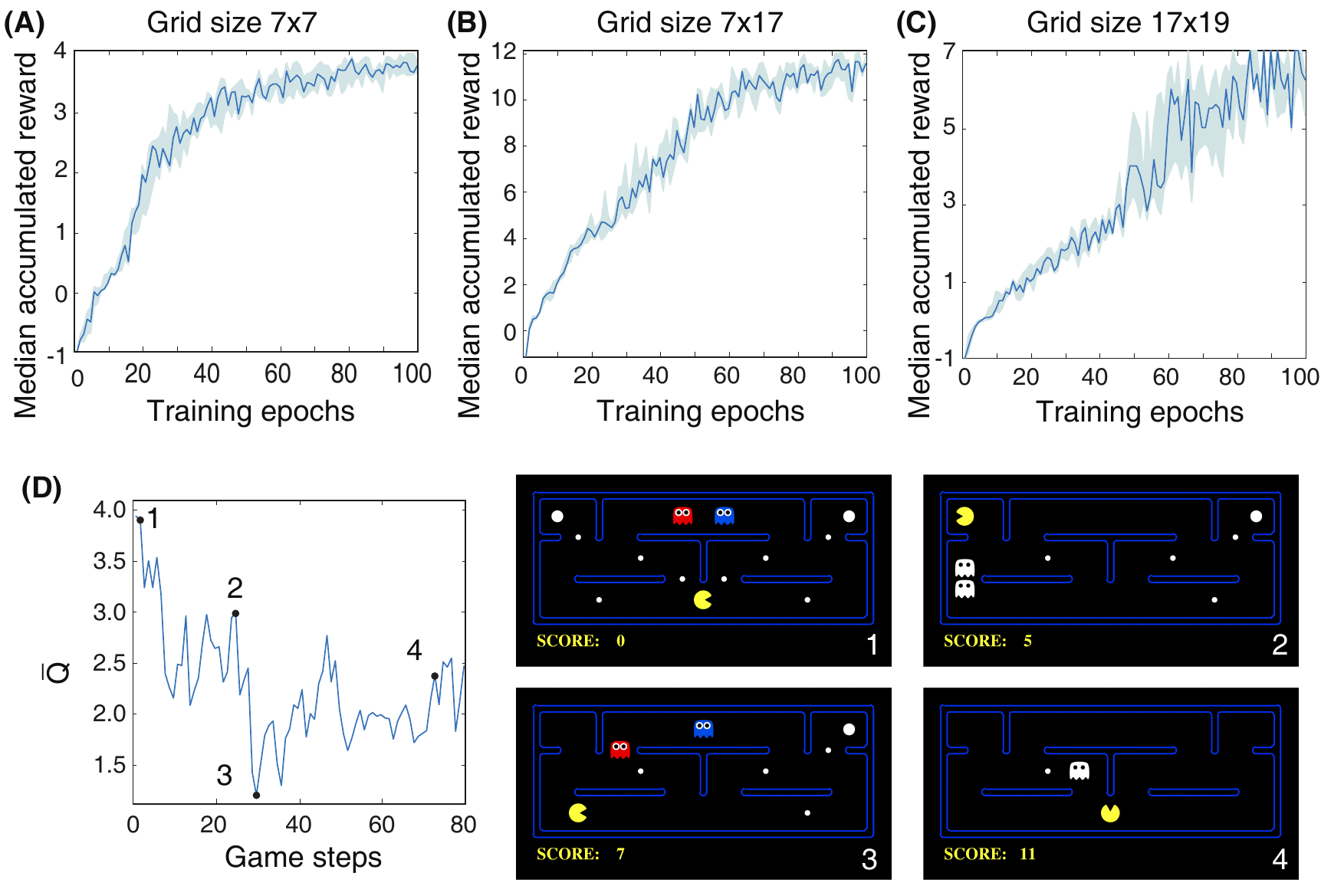}
\end{center}
\caption{\textbf{Median accumulated reward per epoch obtained by training and evaluating the LSM in $3$ different game settings: (A)}  grid size $7\times7$, \textbf{(B)} grid size $7\times17$, and \textbf{(C)} grid size $17\times19$. LSM is initialized and trained with $7$ different initial random seeds. Shaded region represents the $25$-th to $75$-th percentile of the accumulated reward over multiple seeds. \textbf{(D)} The plot on the left shows the predicted state-value function for $80$ continuous Pacman game steps. The four snapshots from the Pacman game shown on the right correspond to game steps designated as 1, 2, 3, and 4, respectively, in the state-value plot.}
\label{fig:result_learning_pacman}
\end{figure}

The LSM configurations and game settings used for Pacman experiments are summarized in \autoref{tab:result_pacmancon}, where each game setting has different degree of complexity with regards to the Pacman grid size and the number of foods, ghosts, and cherries. In the first experiment, we use a $7\times7$ grid with $3$ foods for Pacman to capture and a single ghost to prevent it from achieving its objective. Thus, the maximum possible accumulated reward at the end of a successful game is $4$. \textbf{\autoref{fig:result_learning_pacman}(A)} shows that the median accumulated reward gradually increases with the number of training epochs and converges closer to the maximum possible reward, thereby validating the capability of the liquid to provide useful high-dimensional representation of the environment state necessary for efficient training of the readout weights using the presented methodology. Interestingly, in the second experiment using a larger $7\times17$ grid, we find that the median reward converges to $12$, which is greater than the number of foods available in the grid. This indicates that the LSM does not only learn to capture all the foods; in addition, it also learns to capture the cherry and the scared ghosts, leading to further increase the accumulated reward since consuming the scared ghost results in a unit immediate reward. In the final experiment, we train the LSM to control Pacman in $17\times19$ grid with sparsely dispersed foods. We find that larger grid requires more exploration and training steps for the agent to perform well and achieve the maximum possible reward, resulting in a learning curve that is less steep compared to that obtained for smaller grid sizes in the earlier experiments as shown in \textbf{\autoref{fig:result_learning_pacman}(C)}.

Finally, we plot the average of Q-values produced by the LSM as the Pacman navigates the grid to visualize the correspondence between the learnt Q-values and the enviroment state. As discussed in~\autoref{sec:LSM_training}, each Q-value produced by the LSM provides a measure of how good is a particular action for a give environment state. The Q-value averaged over the set of all possible actions (known as the state-value function) thus indicates the value of being in a certain state. \textbf{\autoref{fig:result_learning_pacman}(D)} illustrates the state-value function while playing the Pacman game in a $7{\times}17$ grid. The predicted state-value starts at a relatively high level because the foods are abundant in the grid and the ghosts are far away from the Pacman (see \textbf{\autoref{fig:result_learning_pacman}(D)-1}). The state-value gradually decreases as the Pacman navigates through the grid and gets closer to the ghosts. The predicted state-value then shoots up after the Pacman consumes cherry and makes the ghosts temporarily consumable (see \textbf{\autoref{fig:result_learning_pacman}(D)-2}), leading to potential additional reward. The predicted state-value drops after the ghosts are reborn (see \textbf{\autoref{fig:result_learning_pacman}(D)-3}). Finally, we observe a slight increase in the state-value towards the end of the game when the Pacman is closer to the last food after it consumes a cherry (see \textbf{\autoref{fig:result_learning_pacman}(D)-4}). It is interesting to note that although the scenario in \textbf{\autoref{fig:result_learning_pacman}(D)-4} is similar to that in \textbf{\autoref{fig:result_learning_pacman}(D)-2}, the state-value is smaller since the expected accumulated reward at this step is at most $3$ assuming that the Pacman can capture both the scared ghost and the last food. On the other hand, in the environment state shown in \textbf{\autoref{fig:result_learning_pacman}(D)-2}, the expected accumulated reward is greater than $3$ since $4$ foods and $2$ scared ghosts are available for the Pacman to capture.

\subsection{Learning to Play Atari Games} \label{sec:finding_atari}

\begin{figure}[!htbp]
\begin{center}
\includegraphics[width=0.90\columnwidth]{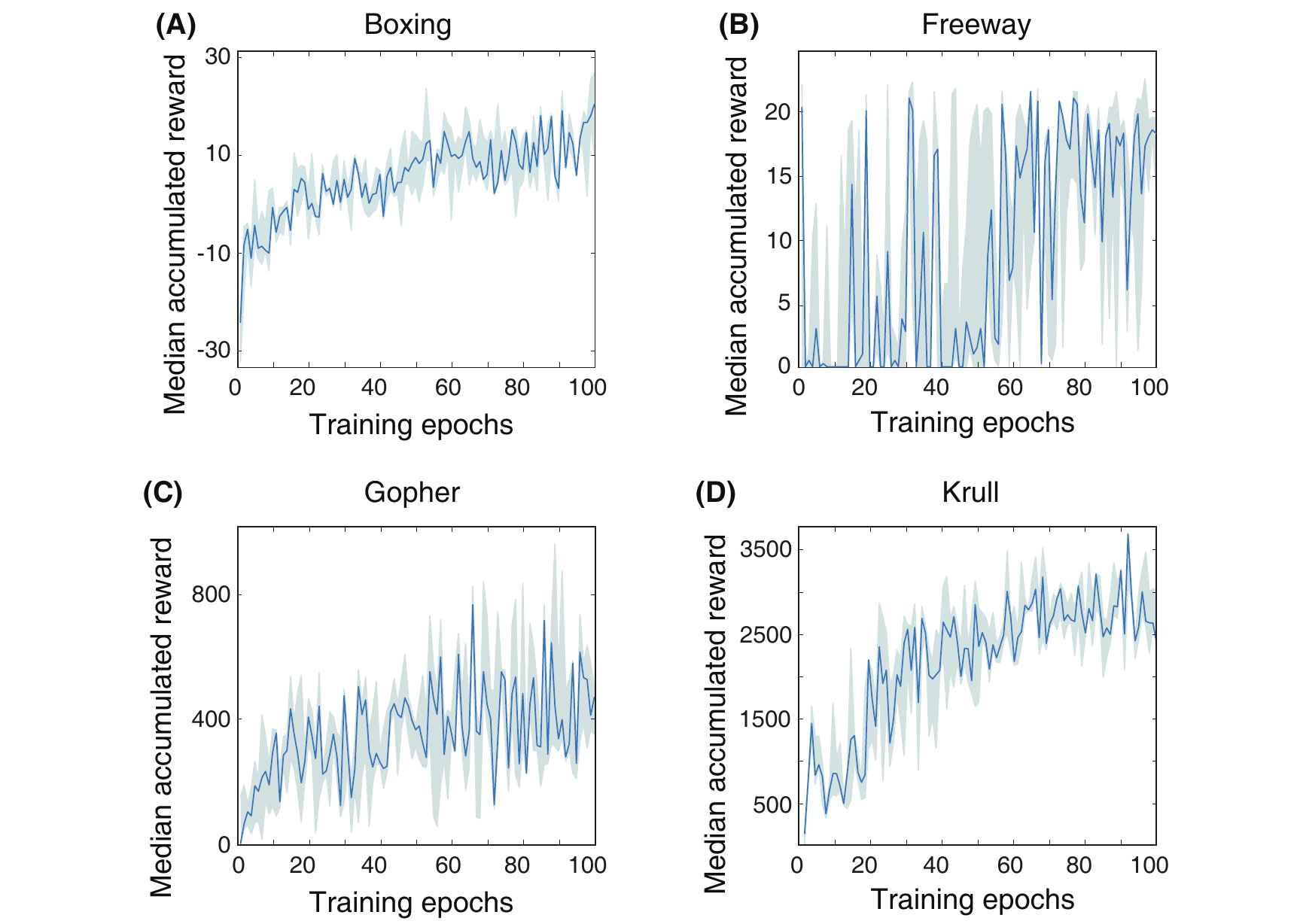}
\end{center}
\caption{\textbf{Median accumulated reward per epoch obtained by training and evaluating the LSM for $4$ selected Atari games: (A)} Boxing, \textbf{(B)} Freeway, \textbf{(C)} Gopher, and \textbf{(D)} Krull. For each game, LSM is initialized and trained with $5$ different initial random seeds. Shaded region represents the $25$-th to $75$-th percentile of the accumulated reward over multiple seeds.}
\label{fig:result_learning_atari}
\end{figure}

Finally, we train the LSM using the presented methodology to play Atari games~\cite{brockman2016openai}, which are widely used to benchmark deep reinforcement learning networks. We arbitrarily select 4 games for evaluation, namely, Boxing, Gopher, Freeway, and Krull. We use the RAM of the Atari machine, which stores 128 bytes of information about an Atari game, as a representation of the environment~\cite{brockman2016openai}. During training, we modified the reward structure of the game by clipping all positive immediate rewards to $1$ and all negative immediate rewards to $-1$. However, we do not clip the immediate reward during testing and measure the actual accumulated reward following \cite{mnih2015human}. For all selected Atari games, we model the agent using an LSM containing $500$ liquid neurons and $128$ hidden neurons. Number of output neurons varies for each game as the number of possible actions is different. The maximum Poisson firing rate for the input neurons is set to $100$~\si{\hertz}. The LSM is trained for $5\times10^{3}$ steps. \textbf{\autoref{fig:result_learning_atari}} illustrates that the LSM learnt the optimal strategies to play Boxing and Krull without any prior knowledge of the rules, leading to high accumulated reward towards the end of the training. However, learning in Gopher and Freeway progresses relatively slow. For detailed evaluation, we compare the median accumulated reward obtained from playing with the trained LSM to the average accumulated reward obtained from playing with random actions for $1\times10^{5}$ steps. We also compare the accumulated reward with that reported for human players in~\cite{mnih2015human}. \autoref{tab:result} shows that the LSM achieves better score than human players on Boxing and Krull while comparable albeit lower score on Freeway and Gopher.

\section{Discussion} \label{sec:discussion}

LSM, an important class of biologically plausible recurrent SNNs, has thus far been primarily demonstrated for pattern (speech/image) recognition \cite{bellec2018long, srinivasan2018spilinc}, gesture recognition \cite{chrolcannon2015learning, panda2018learning}, and sequence generation tasks \cite{panda2017learning, nicola2017supervised, bellec2019biologically} using standard datasets. To the best of our knowledge, our work is the first demonstration of LSMs, trained using Q-learning based methodology, for complex RL tasks like Pacman and Atari games commonly used to evaluate deep reinforcement learning networks. The benefits of the proposed LSM-based RL framework over the state-of-the-art deep learning models are two-fold. First, LSM entails fewer trainable parameters as a result of using fixed input-to-liquid and recurrent-liquid synaptic connections. However, this requires careful initialization of the respective matrices for efficient training of the liquid-to-readout weights as experimentally validated in \autoref{sec:finding}. We note that the stability of LSMs could be further enhanced by training the recurrent weights using localized Spike Timing Dependent Plasticity based learning rules \cite{bi1998synaptic, song2000competitive, diehl2015unsupervised}, which incur lower computational complexity compared to the backpropagation-through-time algorithm \cite{werbos1990backpropagation, bellec2018long} used for training recurrent SNNs. Second, LSMs can be efficiently implemented on event-driven neuromorphic hardware like IBM \textit{TrueNorth} \cite{merolla2014million} or Intel \textit{Loihi} \cite{davies2018loihi}, leading to potentially much improved energy efficiency while achieving comparable performance to deep learning models on the chosen benchmark tasks. Note that the readout layer in the presented LSM needs to be implemented outside the neuromorphic fabric since they are composed of artificial rate-based neurons that are typically not supported in neuromorphic hardware realizations. Alternatively, readout layer composed of spiking neurons could be used that can be trained using spike-based error backpropagation algorithms \cite{panda2016unsupervised, lee2016training, lee2018training, wu2018spatio, jin2018hybrid, bellec2019biologically}. Future works could also explore STDP-based reinforcement learning rules \cite{pfister2006optimal, florian2007reinforcement, farries2007reinforcement, legenstein2008learning} to render the training algorithm amenable for neuromorphic hardware implementations. 

\section{Conclusion} \label{sec:conclusion}

Liquid State Machine (LSM) is a bio-inspired recurrent spiking neural network composed of an input layer sparsely connected to a randomly interlinked liquid of spiking neurons for the real-time processing of spatio-temporal inputs. In this work, we proposed LSMs, trained using the presented Q-learning based methodology, for solving complex Reinforcement Learning (RL) tasks like playing Pacman and Atari that have been hitherto benchmarked for deep reinforcement learning networks. We presented initialization strategies for the fixed input-to-liquid and recurrent-liquid synaptic connectivity matrices and weights to enable the liquid to produce useful high-dimensional representation of the environment state necessary for efficient training of the liquid-to-readout weights. We demonstrated the significance of the inherent capability of the liquid to produce rich representation by training the LSM to successfully balance a cartpole. Our experiments on the Pacman game showed that the LSM learns the optimal strategies for different game settings and grid sizes. Our analyses on a subset of Atari games indicated that the LSM achieves comparable score to that reported for human players in existing works.


\section*{Author Contributions}
Gopalakrishnan Srinivasan and Wachirawit Ponghiran wrote the paper. Wachirawit Ponghiran performed the simulations. All authors helped with developing the concepts, conceiving the experiments, and writing the paper.


\section*{Acknowledgments}
This work was supported in part by the Center for Brain Inspired Computing (C-BRIC), one of the six centers in JUMP, a Semiconductor Research Corporation (SRC) program sponsored by DARPA, by the Semiconductor Research Corporation, the National Science Foundation, Intel Corporation, the DoD Vannevar Bush Fellowship, and by the U.S. Army Research Laboratory and the U.K. Ministry of Defence under Agreement Number W911NF-16-3-0001.

\newpage

\begin{table}[!htbp]
\caption{LSM configuration and game settings for different Pacman experiments reported in this work.} \label{tab:result_pacmancon}
\begin{center}
\renewcommand{\arraystretch}{1.5}
\begin{tabular}{ c c c c c c c c c} 
 \hline
 \textbf{Grid size} & \textbf{Ghost} & \textbf{Food} & \textbf{Cherry} & \textbf{Training steps} & \textbf{Liquid neurons} & \textbf{Hidden neurons}\\
 \hline
 $7{\times}7$ & $1$ & $3$ & $0$ & $5\times10^{5}$ & $500$ & $128$\\
 $7{\times}17$ & $2$ & $6$ & $2$ & $5\times10^{5}$ & $2{,}000$ & $512$\\ 
 $17{\times}19$ & $1$ & $6$ & $0$ & $3\times10^{6}$ & $3{,}000$ & $512$\\ 
 \hline
\end{tabular}
\renewcommand{\arraystretch}{1}
\end{center}
\end{table}

\begin{table}[!htbp]
\caption{Comparison between median accumulated rewarded over multiple random seeds, average accumulated reward from playing with random actions, and accumulated reward from  human game tester reported in \cite{mnih2015human}. Best median accumulated reward over the last $10$ training epochs is reported for each game.} \label{tab:result}
\begin{center}
\renewcommand{\arraystretch}{1.3}
\begin{tabular}{ l c c c } 
 \hline
 \textbf{Game} & \textbf{Human player} & \textbf{Random actions} & \textbf{This work} \\
 \hline
 Boxing & $4.3$ & $0.75$ & $20.2$ \\
 Freeway & $29.6$ & $0.0$ & $19.75$ \\ 
 Gopher & $2{,}321$ & $279.3$ & $611.1$  \\ 
 Krull & $2{,}395$ & $1{,}590$ & $3{,}686$ \\ 
 \hline
\end{tabular}
\renewcommand{\arraystretch}{1}
\end{center}
\end{table}

\begin{table}[!htbp]
\caption{Synaptic weight initialization parameters for the fixed LSM connections.} \label{tab:weight}
\begin{center}
\renewcommand{\arraystretch}{1.5}
\begin{tabular}{ l c } 
 Input-to-liquid connections\\
 \hline
 \textbf{Connection type} & \textbf{Weight} \\
 \hline
 $P{\rightarrow}E$ & $[0,{0.6}]$\\
 \hline
\end{tabular}
\quad
\begin{tabular}{ l c } 
 Recurrent-liquid connections\\
 \hline
 \textbf{Connection type} & \textbf{Weight} \\
 \hline
 $E{\rightarrow}E$ & $[0,{0.05}]$\\ 
 $E{\rightarrow}I$ & $[0,{0.25}]$\\  
 $I{\rightarrow}E$ & $[0,{0.3}]$\\ 
 $I{\rightarrow}I$ & $[0,{0.01}]$\\ 
 \hline
\end{tabular}
\renewcommand{\arraystretch}{1}
\end{center}
\end{table}

\begin{table}[!htbp]
\caption{Leaky-Integrate-and-Fire (LIF) model parameters for the liquid neurons.} \label{tab:neuron_param}
\begin{center}
\renewcommand{\arraystretch}{1.3}
\begin{tabular}{ l c } 
 Excitatory and inhibitory neurons \\
 \hline
 \textbf{Parameter} & \textbf{Value} \\
 \hline
 $V_{rest}$ & $0$ \\
 $V_{reset}$ & $0$ \\
 $V_{thres}$ & $0.5$ \\
 $\tau$ & 20~\si{\milli\second} \\
 ${\tau}_{refrac}$ & 1~\si{\milli\second} \\
 $\Delta t$ (simulation time-step) & 1~\si{\milli\second} \\
 \hline
\end{tabular}
\renewcommand{\arraystretch}{1}
\end{center}
\end{table}

\begin{table}[!htbp]
\caption{Q-learning simulation parameters.} \label{tab:qlearning_param}
\begin{center}
\renewcommand{\arraystretch}{1.3}
\begin{tabular}{ l c c } 
 \hline
 \textbf{Parameter} & & \textbf{Value}\\
 \hline
 Readout weights update frequency & & Once every game-step \\
 Warm up steps before training begins & & $100$ \\
 Batch size for experience replay & & $32$ \\
 Experience replay buffer size & & $1\times10^{6}$ \\
 Discount factor & & $0.95$ \\
 Initial exploration probability during training & & $1$ \\
 Final exploration probability during training (Cartpole) & & $1\times10^{-3}$ \\
 Final exploration probability during training (Pacman \& Atari) & & $1\times10^{-1}$ \\
 Exploration probability during evaluation (Cartpole \& Atari) & & $5\times10^{-2}$ \\
 Exploration probability during evaluation (Pacman) & & $0$ \\
 Learning rate for RMSProp algorithm & & $2\times10^{-4}$ \\
 Term added to denominator for RMSProp algorithm & & $1\times10^{-6}$  \\
 Weight decay for RMSProp algorithm & & $0$ \\
 Smoothing constant for RMSProp algorithm & & $0.99$ \\ 

 \hline
\end{tabular}
\renewcommand{\arraystretch}{1}
\end{center}
\end{table}

\end{document}